\documentclass[letterpaper, 10 pt, conference]{ieeeconf}  % Comment this line out if you need a4paper

\IEEEoverridecommandlockouts                              % This command is only needed if 
\usepackage{graphicx} % for pdf, bitmapped graphics files
\usepackage{amsmath} % assumes amsmath package installed
\usepackage{amssymb}  % assumes amsmath package installed
\usepackage{multirow}
\usepackage{cite}
\usepackage{float}
\usepackage{xcolor}
\title{\LARGE \bf
Anomaly-Driven Approach for Enhanced Prostate Cancer Segmentation
}

\author{Alessia Hu$^{1,2}$ Regina Beets-Tan$^{2,3}$ Lishan Cai$^{2,3}$ Eduardo Pooch$^{2,3}$% <-this % stops a space
\thanks{$^{1}$University of Amsterdam, The Netherlands}%
\thanks{$^{2}$Department of Radiology, The Netherlands Cancer Institute, The Netherlands
}%
\thanks{$^{3}$GROW Research Institute for Oncology and Reproduction, Maastricht University, The Netherlands%
}
}
\newcommand{\methodname}{adU-Net} %annomaly-driven unet?
\begin{document}

\maketitle
\thispagestyle{empty}
\pagestyle{empty}

%%%%%%%%%%%%%%%%%%%%%%%%%%%%%%%%%%%%%%%%%%%%%%%%%%%%%%%%%%%%%%%%%%%%
\begin{abstract}

Magnetic Resonance Imaging (MRI) plays an important role in identifying clinically significant prostate cancer (csPCa), yet automated methods face challenges such as data imbalance, variable tumor sizes, and a lack of annotated data. This study introduces Anomaly-Driven U-Net (\methodname), which incorporates anomaly maps derived from biparametric MRI sequences into a deep learning-based segmentation framework to improve csPCa identification. 
We conduct a comparative analysis of anomaly detection methods and evaluate the integration of anomaly maps into the segmentation pipeline. Anomaly maps, generated using Fixed-Point GAN reconstruction, highlight deviations from normal prostate tissue, guiding the segmentation model to potential cancerous regions. We compare the performance by using the average score, computed as the mean of the AUROC and Average Precision (AP). On the external test set, \methodname{} achieves the best average score of 0.618, outperforming the baseline nnU-Net model (0.605). The results demonstrate that incorporating anomaly detection into segmentation improves generalization and performance, particularly with ADC-based anomaly maps, offering a promising direction for automated csPCa identification.
\newline

% \indent \textit{Clinical relevance}— This is a brief additional statement on why this might be of interest to practicing clinicians. Example: This establishes the anesthetic efficacy of 10\% intraosseous injections with epinephrine to positively influence cardiovascular function.
\end{abstract}

%%%%%%%%%%%%%%%%%%%%%%%%%%%%%%%%%%%%%%%%%%%%%%%%%%%%%%%%%%%%%%%%%%%%%%%%%%%%%%%%
\section{INTRODUCTION}
\label{sec:intro}

Prostate cancer is a significant global health issue, being the second most common cancer in men, with 1,414,000 new cases and over 375,000 deaths in 2020~\cite{prostate_cancer}. 
Tumor segmentation and localization are key for diagnosing cancer, assessing progression, and developing targeted treatment plans, such as radiotherapy. %, which focuses on cancerous tissue while sparing healthy tissue. 
However, manual segmentation remains time-consuming and subjective, with high interobserver variability~\cite{gillespie2020deep}. 
Deep learning methods present a promising solution by automating tumor segmentation, reducing interobserver variability, and potentially enhancing diagnostic accuracy.

% speak a bit on bpMRI
Segmenting clinically significant prostate cancer (csPCa) presents several challenges, primarily due to the complex nature of prostate tissue, lack of training data, poor generalization, and high variability in lesion appearances. Larger, more diverse datasets are needed, but the class imbalance in lesion detection datasets can still limit model performance and generalization.
One solution to tackle data imbalance in medical imaging is to integrate anomaly detection techniques into segmentation models. Multiparametric magnetic resonance imaging (mpMRI) is a well-established diagnostic tool for PCa, but biparametric MRI (bpMRI) is gaining popularity due to its reduced scan time, lower costs, and comparable diagnostic accuracy to mpMRI.

Anomaly maps highlight irregular areas in an image in contrast to the "healthy" portion of the data. In medical imaging, these deviations can be used to detect and visualize areas of concern, such as tumors or other pathological changes~\cite{baur2021autoencoders}.
This approach primarily trains on healthy images, therefore bypassing the issue of limited annotated lesions. 

This study presents two main contributions. First, we provide a comparative evaluation of anomaly detection methods for prostate bpMRI. Second, we propose anomaly-driven U-Net (adU-Net) and demonstrate that anomaly maps can enhance segmentation by guiding the model’s focus towards possible abnormal regions.

\section{METHODOLOGY}
\label{sec:methodology}

\begin{figure*}[t]
    \centering
    \begin{minipage}[b]{0.8\linewidth}
        \centering
        \includegraphics[width=\textwidth]{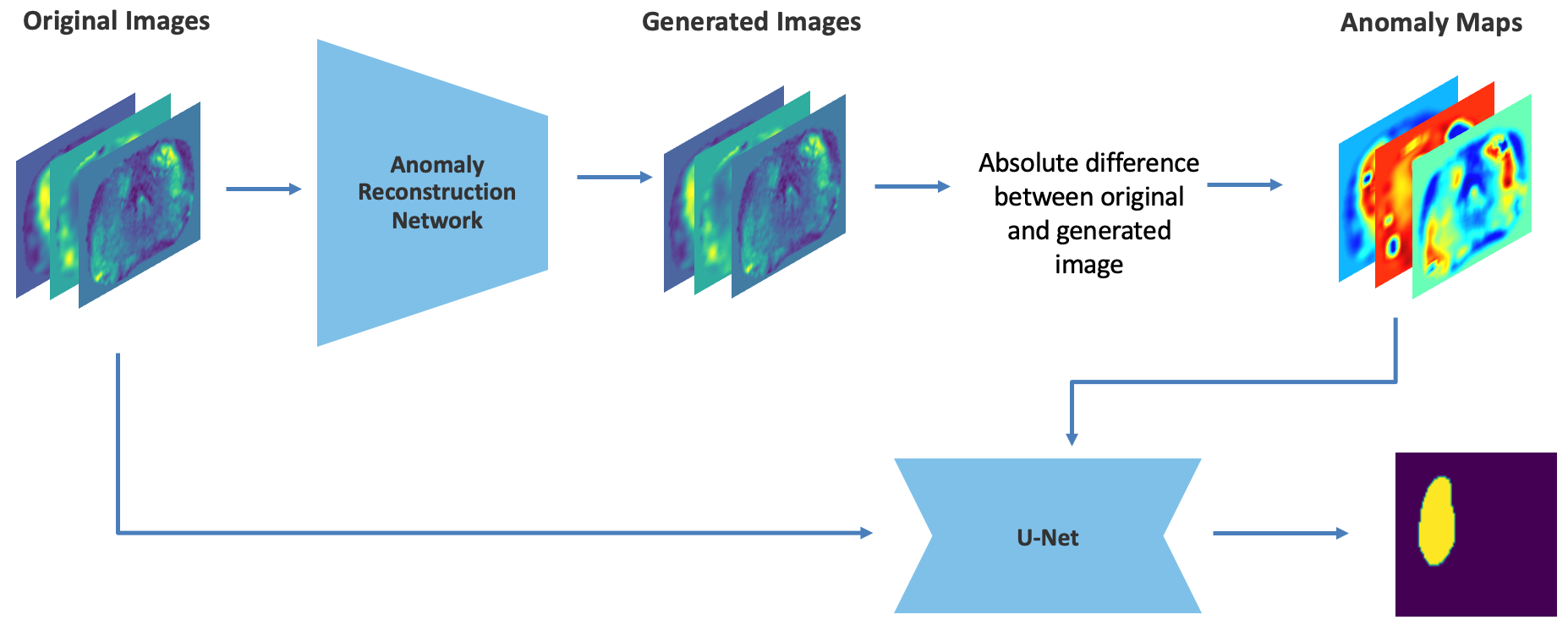}
    \end{minipage}
    \caption{Illustration of the proposed approach. The original MRI image is fed into an anomaly reconstruction network, which generates reconstructed images. By calculating the absolute difference between the original and reconstructed images, anomaly maps are obtained. These anomaly maps, along with the original images, are then fed as input to the adU-Net model, which generates a segmentation map for csPCa detection.}
    \label{fig:overview_proposed_approach}
\end{figure*}

Our proposed method \methodname{} aims to exploit the information within the generated anomaly maps to improve network performance. Specifically, \methodname{} integrates these maps as additional channels when passing the input image through a segmentation network—in this study, nnU-Net \cite{nnunet}.
This approach helps guide the network's focus toward regions of interest, particularly in identifying csPCa. The integration is achieved by concatenating the anomaly map with the input. An overview of the pipeline is shown in Figure \ref{fig:overview_proposed_approach}.

First, we generate anomaly maps $a$ by providing an input image $x$ to a GAN or diffusion model $G(\cdot)$, which reconstructs $x$ into a pseudo-healthy image $G(x)$. The anomaly $a$, indicating potential cancerous regions, is calculated by taking the absolute difference between the input image and the generated output (see Equation \ref{eq:anomaly_map}). To reduce noise, this difference map is smoothed using a blurring filter $\Phi$ with a kernel size of 8.

\begin{equation}
a = \Phi ( | x - G(x) |)
\label{eq:anomaly_map}
\end{equation}

For segmentation purposes, some additional processing steps are required. First, the background of the generated anomaly maps is masked out to prevent the introduction of noise from irrelevant regions. During training, it is also necessary to resample slices containing relevant information, as some slices may not include the prostate. This resampling aligns these slices with the original image’s slice order. Finally, to ensure consistency with the csPCa segmentation training data, the images used as additional input channels must be resampled to match the original domain, including origin, direction, and spacing.

\subsection{Anomaly Detection Architectures}
In this study, we employed four reconstruction methods to generate anomaly maps: a Dense Autoencoder, a Spatial Autoencoder, a Diffusion Model, and a Fixed-Point GAN.

We initially employed Dense and Spatial Autoencoders, adapted from the architectures proposed by Baur et al.~\cite{baur2021autoencoders}, to establish a foundational approach for image reconstruction. These simplified models enabled a preliminary investigation into anomaly detection by evaluating their capacity to accurately reconstruct healthy images.

For the Diffusion Model, we followed the approach by Wolleb et al.~\cite{wolleb2022diffusion}, where input images undergo iterative noise addition, and a U-Net model is trained to reverse this process by removing the noise. This approach is guided by a classifier trained on healthy images, ensuring that the reconstructed image tends toward the "healthy" class, effectively allowing for anomaly detection by highlighting deviations.

The Fixed-Point GAN~\cite{fp_gan} further enhances this reconstruction approach by translating diseased images into their healthy counterparts. The GAN achieves this through two phases: in the same-domain translation, the model ensures identity mapping to maintain key attributes, while the cross-domain translation focuses on converting diseased images into healthy ones. Adversarial and cycle-consistency losses ensure the accuracy and realism of these transformations.

\section{EXPERIMENTS}
\label{sec:experiments}
This section outlines the data utilized in this study, the preprocessing steps taken to ensure data quality and consistency, and the experimental settings for the various networks. All frameworks were implemented using PyTorch. The experiments were conducted on an NVIDIA GeForce RTX 2080 Ti GPU with 11GB of memory.

\subsection{Data}
The first dataset used in this work is the PI-CAI challenge dataset~\cite{saha2024artificial}, which supports the training and validation of segmentation and anomaly detection methods. For external testing, the Prostate158 dataset~\cite{adams2022prostate158} was used. Both datasets are publicly available

\begin{figure*}[t]
    \centering
    \includegraphics[width=0.9\textwidth]{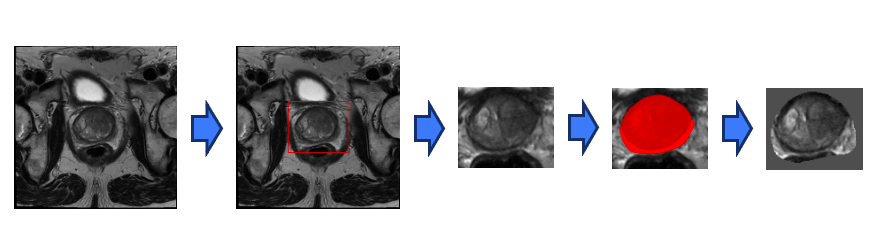}
    \caption{Overview of the preprocessing pipeline used to improve prostate reconstruction. The pipeline begins with the original image, followed by the identification of a bounding box around the prostate. The image is then cropped to this region of interest, reducing irrelevant background information. Finally, a prostate mask is applied to isolate the prostate and remove surrounding structures, ensuring that the reconstruction models focus on the most relevant anatomical region.}
    \label{fig:preprocess}
\end{figure*}

The PI-CAI dataset includes 1500 bpMRI images with three modalities—T2W, DWI, and ADC—alongside a binary segmentation map indicating clinically significant Prostate Cancer. Images were preprocessed using the nnU-Net~\cite{nnunet} framework. For anomaly reconstruction tasks, only patients without visible abnormalities were included, randomly selecting a subset of 100 patients to ensure efficient training while preserving representativeness. To improve reconstruction quality and reduce interference from surrounding tissues, the images were first cropped using predicted segmentation masks of the peripheral zone (PZ) and transition zone (TZ). These masks were generated by an nnU-Net segmentation model trained on the ProstateX dataset \cite{armato2018prostatex}, which provides ground-truth labels for PZ and TZ. By cropping around these regions, irrelevant tissues and structures outside the areas of interest were removed. Additionally, the mask was used to eliminate the background, leaving only the prostate tissue. This approach significantly reduces noise from surrounding tissues. The pre-processing pipeline is shown in Figure \ref{fig:preprocess}.

The Prostate158 dataset consists of 158 annotated biparametric 3T prostate MRIs, including T2W and DW images, ADC maps, and pixel-wise segmentations of the central gland, PZ, and csPCa lesions. The anatomical segmentations were used similarly to the PI-CAI dataset for cropping and masking images. The segmentation masks of the anatomical zones were predicted using the publicly available baseline model weights. 

\begin{figure*}[t]
  \centering
  \includegraphics[width=0.9\textwidth]{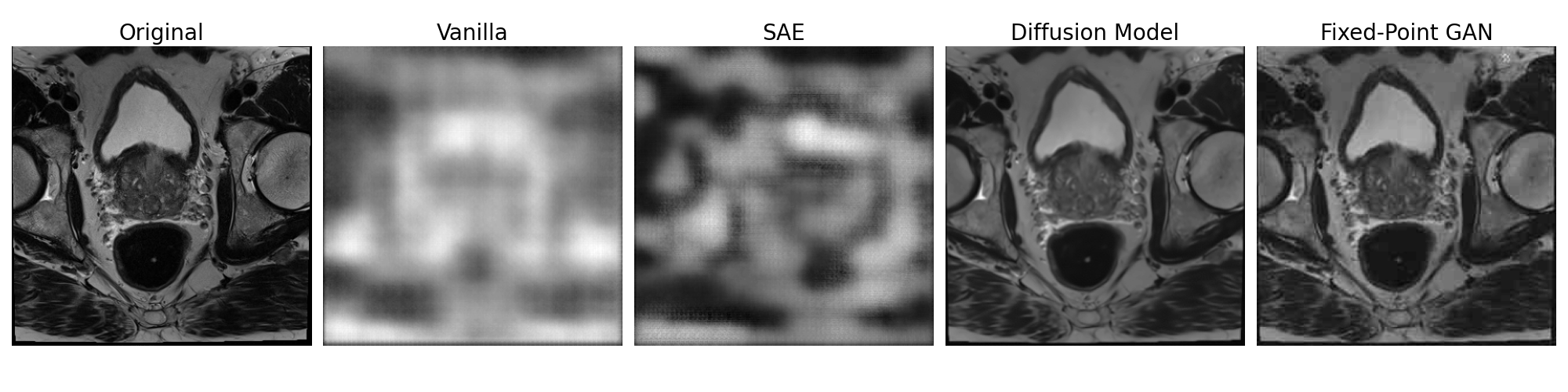}
  \caption{Comparison of reconstruction performance across different architectures. The vanilla and spatial autoencoders struggle with fine structural details, while the Diffusion Model and Fixed-Point GAN generate more accurate and high-resolution reconstructions.}
  \label{fig:recon}
\end{figure*}

\subsection{Anomaly Dectection}
This section details the experiments conducted to evaluate various reconstruction architectures for anomaly detection. 
Initially, experiments were conducted on full images; however, the results demonstrated poor performance, likely due to the inclusion of irrelevant background information and the varying scales of anomalies present in the full images. To mitigate these issues and improve the model's accuracy, the images were cropped to focus on the regions of interest where anomalies were more likely to occur. This adjustment aimed to reduce noise and enhance the model's ability to detect and reconstruct anomalies effectively.

\textbf{Dense and Spatial AE}
The networks adapted from Baur et al.~\cite{baur2021autoencoders} underwent a series of experiments to identify the optimal hyperparameters. The final configuration for these adapted networks incorporated the L2 loss function and implemented a learning rate scheduler with a step size of 30 and an initial learning rate of 0.1. 

\textbf{Diffusion Model}
Training the diffusion model involved two key components: the classifier and the Denoising Diffusion Probabilistic Model (DDPM). For the DDPM, there was no data augmentation, and the model used 1,000 sampling steps with a linear noise schedule over 400,000 training iterations. The classifier was optimized using the Adam optimizer with a learning rate of 1e-4, a batch size of 10, 128 channels in the first layer, and an attention head resolution of 16, trained over 150,000 iterations. %These experiments were conducted on an NVIDIA GeForce RTX 2080 Ti GPU, with training times of approximately 1 day for the classifier and 3 days for the diffusion model.

\textbf{Fixed-Point GAN}
The training configuration for the Fixed-Point GAN involved several parameters. The generator (G) and discriminator (D) networks were initialized with 64 convolutional filters in their first layers. The generator included six residual blocks, while the discriminator consisted of six strided convolutional layers. The training was performed with a batch size of 16 over 200,000 iterations, using a learning rate of 0.0001 for both G and D, with a decay starting at 100,000 iterations.

\section{RESULTS}
\subsection{Anomaly Detection}

Image reconstruction quality was assessed through Peak Signal-to-Noise Ratio (PSNR) and Structural Similarity Index Measure (SSIM). PSNR measures how much noise is present in a reconstructed image relative to the original, with higher values indicating less distortion. Meanwhile, SSIM assesses structural and contrast similarities between the original and reconstructed images, with values closer to 1 signifying higher preservation of fine details. These metrics are particularly important for prostate MRI, where subtle anatomical variations are key for identifying csPCa lesions.
The qualitative outcomes of these architectures are presented in Figure \ref{fig:recon}. The figure shows the models' capabilities in reconstructing input images and highlights the performance differences between the architectures.
The vanilla and spatial autoencoder failed to capture fine structures and produce high-resolution reconstructions, limiting their ability to highlight cancerous regions. Although these architectures performed well on brain MRI, their application to prostate MRI exposed limitations, largely due to differences in tissue characteristics and the increased movement in prostate imaging compared to brain MRI. These factors contribute to challenges in capturing and reconstructing anomalies specific to the prostate. In contrast, the Diffusion Model and Fixed-Point GAN produced more accurate, detailed reconstructions.

\begin{figure}[h]
  \centering
  \includegraphics[width=0.46\textwidth]{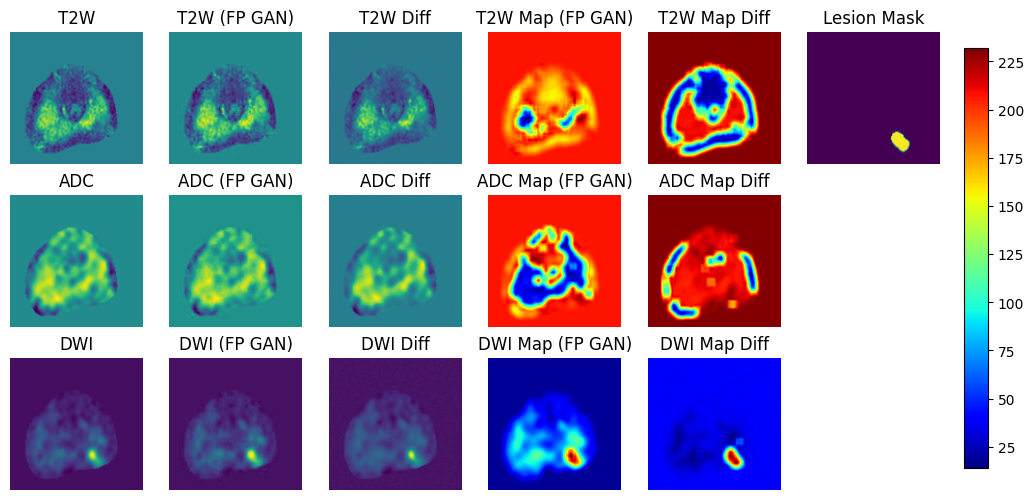}

  \includegraphics[width=0.46\textwidth]{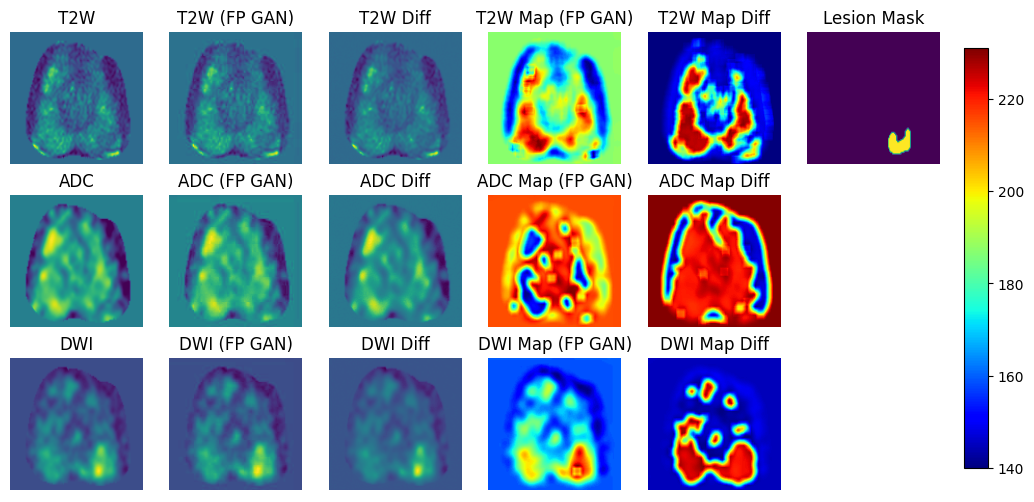}

  \includegraphics[width=0.46\textwidth]{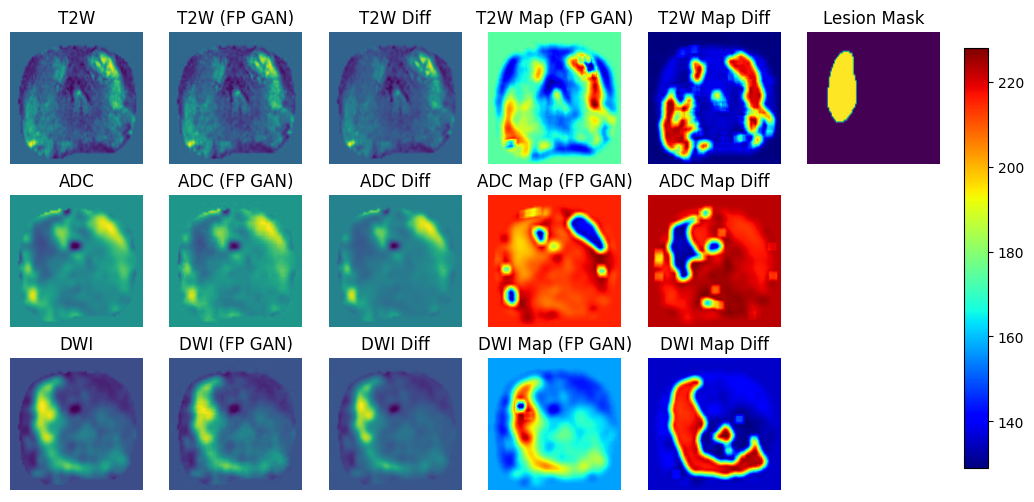}

  \includegraphics[width=0.46\textwidth]{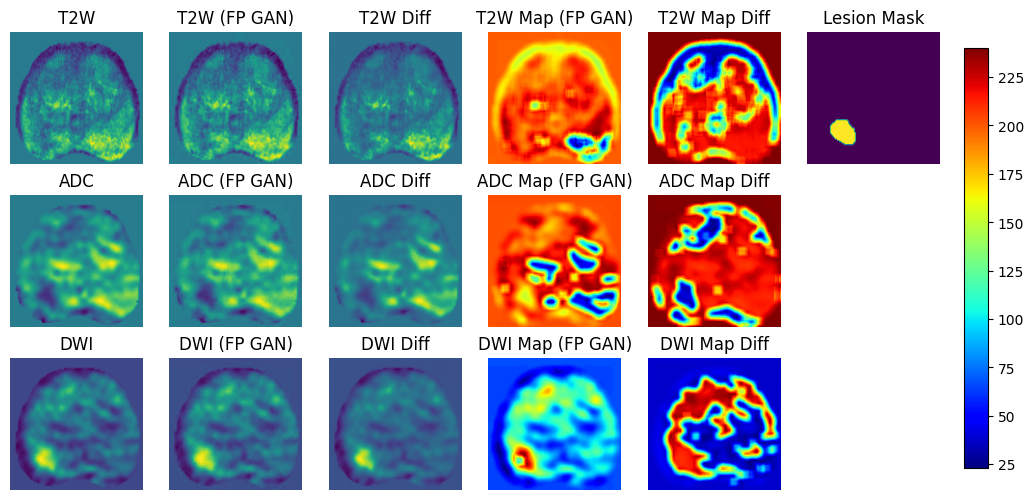}

  \caption{Visualization of anomaly maps for four patients. The first column shows the original MRI sequences (T2W, ADC, and DWI). The second and third columns display reconstructions from FP-GAN and the diffusion model. The fourth and fifth columns show the corresponding anomaly maps, while the final column presents the lesion masks highlighting clinically annotated csPCa regions.}
  \label{fig:all_figs}
\end{figure}

The quantitative results in Table \ref{tab:results_recon} confirmed the qualitative findings. FP GAN consistently outperformed other models, achieving the highest SSIM and PSNR values across all modalities, particularly in cropped images where fine detail preservation is crucial. In T2W, FP GAN reached SSIM 0.939 and PSNR 30.532, outperforming the Diffusion Model (SSIM 0.8547, PSNR 24.029), demonstrating superior detail retention.
The "Test Results" section in Table \ref{tab:results_recon} presents model performance on an unseen test set, ensuring that the findings generalize beyond the validation set. FP GAN maintained strong SSIM and PSNR scores in T2W (0.9480, 28.079) and ADC (0.8730, 28.528), indicating robust generalization. However, DWI performance dropped (SSIM 0.6916), highlighting challenges in reconstructing noisy diffusion-weighted images. These results confirmed that FP GAN’s ability to handle complex tissue structures extends beyond the training data, reinforcing its suitability for prostate MRI anomaly detection.
Figure \ref{fig:all_figs} presents 4 slices from different patients, showing the original images, reconstructions by FP GAN and the diffusion model, their anomaly maps, and corresponding lesion masks. While the Diffusion Model performed well, it shows more pronounced differences, indicating room for improvement in fine detail preservation. Overall, FP GAN provided better localization and image quality.

The findings in Table \ref{tab:results_recon} show that FP GAN outperformed the diffusion model across all metrics and modalities (T2W, ADC, DWI). Based on these results, FP GAN-generated anomaly maps were used to enhance the segmentation network. FP GAN's superior performance on prostate MRI data is due to its ability to capture subtle anatomical variations and specific prostate features, crucial for accurate anomaly detection. The use of fixed-point iterations stabilizes training and enhances anomaly localization. Additionally, FP GAN effectively handles complex tissue structures and lesion variations in prostate MRI, making it more robust than the diffusion model in this context.

\begin{table}[h]
\centering
\small
\caption{Quantitative results of image reconstruction from different architectures. Initially, all networks were used to reconstruct full images. The best performing architectures, Diffusion and FP-GAN, were then selected for cropped image reconstructions, with FP-GAN being the only architecture used for the final test on cropped images due to its superior performance.}
\begin{tabular}{|c|c|c|c|}
\hline
\textbf{Method} & \textbf{Image Modality} & \textbf{SSIM} & \textbf{PSNR (dB)} \\ \hline
\multirow{3}{*}{\centering Vanilla AE} & T2W & 0.1885 & 11.8812 \\ 
 & ADC & 0.1692 & 14.0192 \\ 
 & DWI & 0.2475 & 14.0217 \\ \hline
\multirow{3}{*}{\centering Spatial AE} & T2W & 0.2401 & 14.1857 \\ 
 & ADC & 0.2478 & 17.1155 \\ 
 & DWI & 0.2891 & 15.5884 \\ \hline
\multirow{3}{*}{\centering Diffusion Model} & T2W & 0.7456 & 23.8554 \\ 
 & ADC & 0.7647 & 25.0425 \\ 
 & DWI & 0.7673 & 27.4743 \\ \hline
\multirow{3}{*}{\centering FP GAN} & T2W & 0.7902 & 21.9702 \\ 
 & ADC & 0.7515 & 21.666 \\ 
 & DWI & 0.7445 & 24.9116 \\ \hline
\multicolumn{4}{|c|}{\textbf{Cropped Images Comparison}} \\ \hline
\multirow{3}{*}{\centering Diffusion Model} & T2W & 0.8547 & 24.029 \\ 
 & ADC & 0.8895 & 21.841 \\ 
 & DWI & 0.8562 & 22.005 \\ \hline
\multirow{3}{*}{\centering FP GAN} & T2W & 0.939 & 30.532 \\ 
 & ADC & 0.957 & \textbf{31.475} \\ 
 & DWI & \textbf{0.961} & 31.342 \\ \hline
\multicolumn{4}{|c|}{\textbf{Test Results}} \\ \hline
\multirow{3}{*}{\centering FP GAN} & T2W & 0.9480 & 28.079 \\ 
 & ADC & 0.8730 & 28.528 \\ 
 & DWI & 0.6916 & 27.8080 \\ \hline
\end{tabular}

\label{tab:results_recon}
\end{table}

\subsection{Segmentation}

Segmentation evaluation is conducted using the Area Under the Receiver Operating Characteristic curve (AUROC) for patient-level diagnosis, Average Precision (AP) for lesion-level diagnosis, and the overall average score, following the evaluation metrics of the PI-CAI challenge \cite{saha2024artificial}. While the Dice score is a common segmentation evaluation metric, it is not fully appropriate in this context. Due to the inherent class imbalance in csPCa segmentation – where many MRI scans lack lesions – the Dice score is less informative, as it does not account for correctly predicting lesion absence.

Table \ref{tab:results} shows that incorporating additional MRI sequences improves segmentation performance for csPCa detection. On the validation set, nnU-Net achieves the highest AUROC (0.7985), establishing a strong baseline. However, \methodname{} with ADC anomaly maps attains the best AP (0.4437), underscoring ADC’s importance in lesion detection. While \methodname{} (all bpMRI) does not surpass nnU-Net in AUROC, it achieves a comparable overall average score (0.6150 vs. 0.6157), indicating that combining sequences maintains strong performance. Notably, \methodname{} with DWI alone performs the worst (AUROC: 0.6584, AP: 0.2702), likely due to higher noise levels in DWI images.

On the external test set, \methodname{} (ADC) reaches the highest AUROC (0.8026), demonstrating its robustness across datasets. Meanwhile, \methodname{} (all bpMRI) attains the best overall average score (0.6181), reinforcing the benefit of leveraging multi-modal information. DWI alone continues to underperform but contributes positively when combined with other sequences, likely due to complementary information compensating for its inherent noise.

Overall, these findings highlight the importance of ADC in detecting csPCa and suggest that integrating multiple sequences improves segmentation robustness over single-sequence models. While nnU-Net remains a strong baseline, multi-modality approaches, particularly \methodname{} (all bpMRI), show potential for further improvement.

While there are some performance variations between the validation and external test sets, these differences can be attributed to dataset shifts, including variations in image acquisition, scanner differences, and patient demographics. The overall trends remain consistent, with ADC-based methods performing well and multi-modal approaches offering robust generalization. Notably, the proposed method demonstrates superior performance over the baseline on the external test set, highlighting its potential for improving the generalization of segmentation models, particularly with ADC-based and multi-modal approaches. 

\begin{table}[h]
\centering
\small
\caption{Comparison of model performance on the validation and external test sets for csPCa segmentation. All models receive the three bpMRI sequences as input. The nnU-Net baseline serves as a reference, while different versions of \methodname incorporate specific anomaly maps (indicated in parentheses) into the input.}
\label{tab:results}
\begin{tabular}{|l|c|c|c|}
\hline
\multicolumn{4}{|c|}{\textbf{Validation set}} \\ \hline
Model & AUROC & AP & Average \\
\hline
nnU-Net baseline & \textbf{0.7985} & 0.4329 & \textbf{0.6157} \\
\hline 
\methodname (T2W) & 0.7622  & 0.4106 & 0.5864\\
\methodname (ADC) & 0.7834  & \textbf{0.4437} & 0.6135\\
\methodname (DWI) & 0.6584 & 0.2702 & 0.4643\\
\methodname (all bpMRI) & 0.7857  & 0.4436  & 0.6150 \\

\hline
\multicolumn{4}{|c|}{\textbf{External test set}} \\ \hline
Model & AUROC & AP & Average \\
\hline

nnU-Net baseline & 0.7737 & 0.436  & 0.6049 \\
\hline
\methodname (T2W)  & 0.7975  & 0.4204 & 0.6090\\

\methodname (ADC)  & \textbf{0.8026}  & 0.4306 & 0.6166\\
\methodname (DWI)  & 0.7533  & 0.4217 & 0.5875\\
\methodname (all bpMRI) & 0.7899 & \textbf{0.4463} & \textbf{0.6181} \\
\hline
\end{tabular}
\end{table}

\section{CONCLUSION}
This study demonstrates the potential of integrating anomaly maps alongside multi-parametric MRI data within a U-Net-based architecture to improve the segmentation performance for clinically significant prostate cancer (csPCa). While relying on established architectures like U-Net, our proposed method, \methodname, achieves improved performance in detecting csPCa compared to a standard nnU-Net baseline, particularly highlighting its value on an independent external test set. This suggests that the explicit incorporation of anomaly information, representing deviations from healthy tissue, enhances the model's ability to discern subtle differences between cancerous and non-cancerous regions. However, further investigation into the variance and characteristics of these anomaly maps is required to fully understand their contribution. Future work should explore advanced multi-modal fusion strategies to better capture complementary information from different MRI sequences, as well as refinement of anomaly detection techniques and application of this approach to other complex medical imaging segmentation tasks. 

\addtolength{\textheight}{-12cm}   % This command serves to balance the column lengths

%%%%%%%%%%%%%%%%%%%%%%%%%%%%%%%%%%%%%%%%%%%%%%%%%%%%%%%%%%%%%%%%%%%%%%%%%%%%%%%%

%%%%%%%%%%%%%%%%%%%%%%%%%%%%%%%%%%%%%%%%%%%%%%%%%%%%%%%%%%%%%%%%%%%%%%%%%%%%%%%%



\section*{ACKNOWLEDGMENTS}
The authors would like to acknowledge the Research High Performance Computing (RHPC) facility of the Netherlands Cancer Institute (NKI).

%%%%%%%%%%%%%%%%%%%%%%%%%%%%%%%%%%%%%%%%%%%%%%%%%%%%%%%%%%%%%%%%%%%%%%%%%%%%%%%%

% \begin{thebibliography}{99}
% \bibliographystyle{IEEEbib}
\bibliographystyle{plain}
\bibliography{refs}

% \end{thebibliography}

\end{document}